\PassOptionsToPackage{table,xcdraw}{xcolor}
\documentclass[letterpaper, 10 pt, conference]{ieeeconf}

\IEEEoverridecommandlockouts                              
\DeclareFontFamily{U}{mathb}{\hyphenchar\font45}
\DeclareFontShape{U}{mathb}{m}{n}{
<-6> mathb5 <6-7> mathb6 <7-8> mathb7 <8-9> mathb8
<9-10> mathb9 <10-12> mathb10 <12-> mathb12}{}
\DeclareSymbolFont{mathb}{U}{mathb}{m}{n}
\DeclareMathSymbol{\llcurly}{\mathrel}{mathb}{"CE}
\DeclareMathSymbol{\ggcurly}{\mathrel}{mathb}{"CF}

\usepackage{amsmath,amssymb}
\usepackage{graphicx}
\usepackage{orcidlink}
\usepackage{hyperref}
\usepackage{cleveref}
\usepackage{subcaption}
\usepackage{booktabs}
\usepackage{xcolor}
\definecolor{grn}{HTML}{4EA72E}
\usepackage{multirow}

\makeatletter
\let\NAT@parse\undefined
\makeatother
\usepackage[numbers]{natbib}

\crefname{figure}{Fig.}{Figs.}
\Crefname{figure}{Figure}{Figures}
\crefname{section}{Sect.}{Sects.}
\Crefname{section}{Section}{Sections}

\DeclareMathOperator{\argmax}{argmax}

\title{\LARGE \bf%
How Hard is Snow? A Paired Domain Adaptation Dataset\\for Clear and Snowy Weather: CADC+}

\author{
Mei Qi Tang\,\orcidlink{0009-0000-9867-6364}, Sean Sedwards\,\orcidlink{0000-0002-2903-0823}, Chengjie Huang\,\orcidlink{0009-0002-6693-0828}, Krzysztof Czarnecki\,\orcidlink{0000-0003-1642-1101}\\
\texttt{\small\{mqtang,sean.sedwards,c.huang,k2czarne\}@uwaterloo.ca}\\
University of Waterloo, Canada}

\begin{document}
\bstctlcite{IEEEexample:BSTcontrol} 
\maketitle
\thispagestyle{empty}
\pagestyle{empty}

\begin{abstract}

Evaluating the impact of snowfall on 3D object detection requires a dataset with sufficient labelled data from both snowy and clear weather conditions, ideally captured in the same driving environment. Current datasets with LiDAR point clouds either do not provide enough labelled data in both domains, or rely on de-snowing methods to generate synthetic clear weather. Synthetic data often lacks realism and introduces an additional domain shift that confounds accurate evaluations. To address these challenges, we present CADC+, the first paired weather domain adaptation dataset for autonomous driving in winter conditions.\footnote{The dataset can be downloaded at \href{https://uwaterloo.ca/waterloo-intelligent-systems-engineering-lab/cadc-plus}{https://uwaterloo.ca/waterloo-intelligent-systems-engineering-lab/cadc-plus}.} CADC+ extends the Canadian Adverse Driving Conditions (CADC) dataset using clear weather data that was recorded on the same roads and in the same period as CADC. To create CADC+, we pair each CADC sequence with a clear weather sequence that matches the snowy sequence as closely as possible. CADC+ thus minimizes the domain shift resulting from factors unrelated to the presence of snow. We also present preliminary results using CADC+ to evaluate the effect of snow on 3D object detection performance. We observe that snow introduces a combination of aleatoric and epistemic uncertainties, acting as both noise and a distinct data domain. 


\end{abstract}

\section{Introduction}

LiDAR (light detection and ranging) sensors generate accurate 3D maps of the environment by measuring distances using laser pulses, making them the gold standard for 3D object detection in autonomous driving. However, various studies show that adverse weather conditions affect the laser pulses that these devices emit: hydrometeor particles, such as snowflakes, can affect LiDAR scans due to signal attenuation and backscattering, resulting in spurious points, missing signal returns, and points with weaker intensity~\cite{Rasshofer2011, Jokela2019, Kutila2020}. Despite this knowledge, research on the impact of snow on 3D object detectors remains limited. 

\begin{figure}[t]
    \centering
    \begin{subfigure}[b]{0.48\linewidth}
        \centering
        \includegraphics[width=\linewidth]{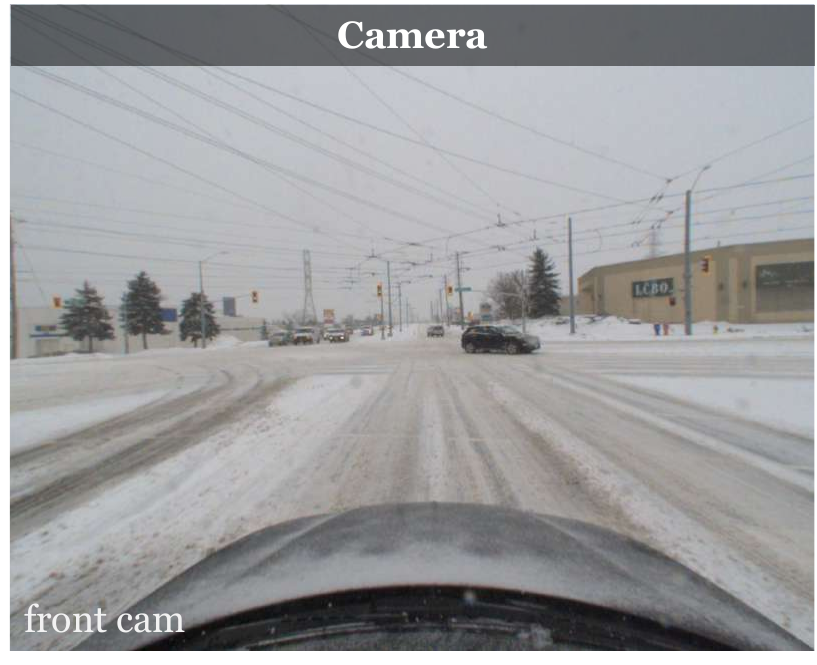}
        \label{fig:cadc_example_1}
    \end{subfigure}
    \hfill
    \vspace{-1.3em}
    \begin{subfigure}[b]{0.48\linewidth}
        \centering
        \includegraphics[width=\linewidth]{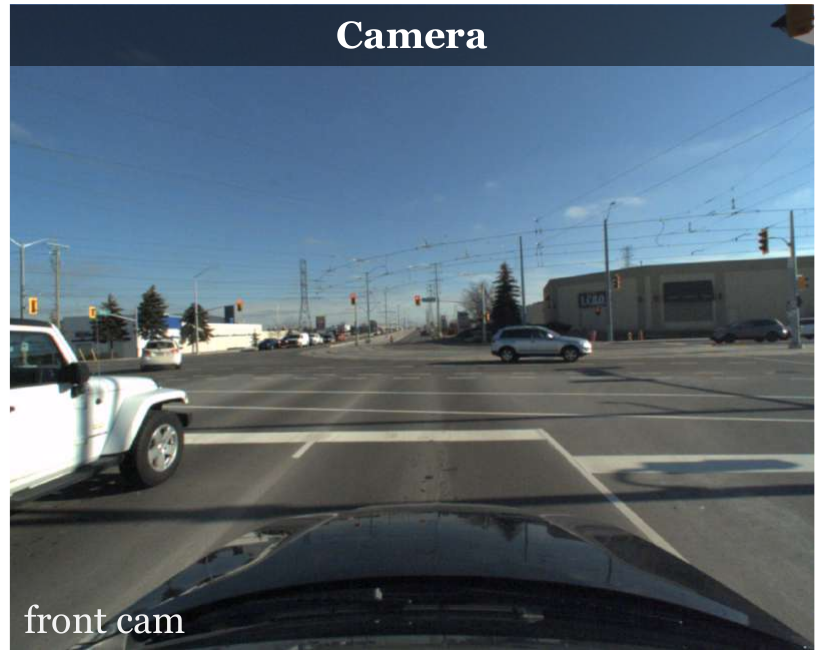}
        \label{fig:clear_example_1}
    \end{subfigure}
    
    \begin{subfigure}[b]{0.48\linewidth}
        \centering
        \includegraphics[width=\linewidth]{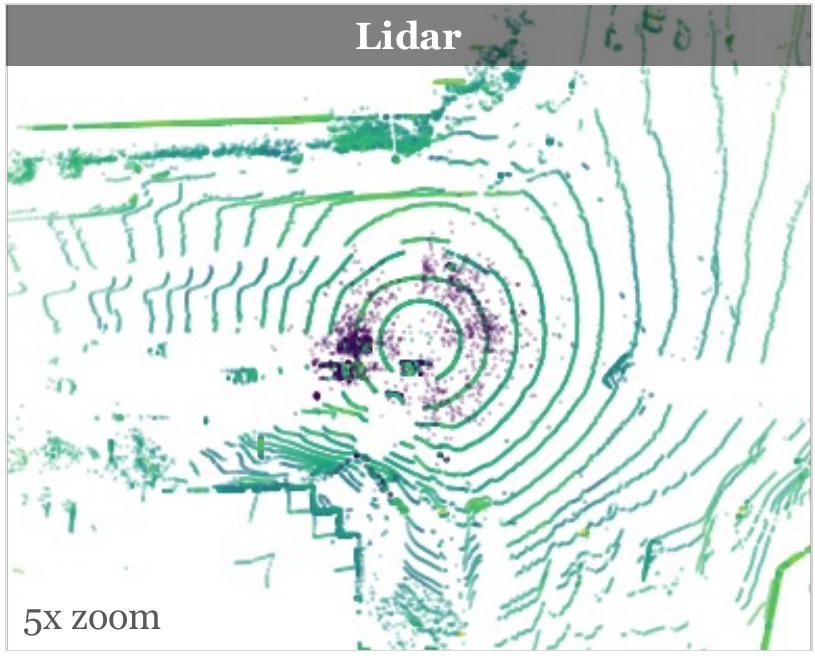}
        \caption{CADC}
        \label{fig:cadc_example_2}
    \end{subfigure}
    \hfill
    \begin{subfigure}[b]{0.48\linewidth}
        \centering
        \includegraphics[width=\linewidth]{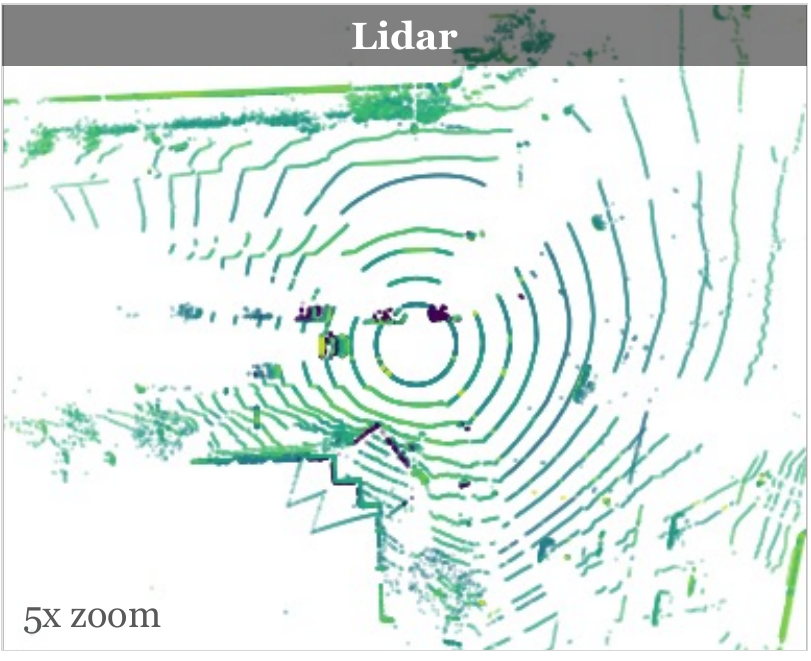}
        \caption{CADC-clear}
        \label{fig:clear_example_2}
    \end{subfigure}
    
    \caption{Example of corresponding data frames in paired sequences from CADC+.}
    \label{fig:cadc_and_clear_examples}
\end{figure}

To evaluate the impact of snowy conditions, 3D object detectors are ideally evaluated on weather-paired datasets, where snow is the \emph{only} varying factor, with otherwise identical scenes. However, creating such datasets is impractical. Existing snowy weather LiDAR datasets make no attempt to pair sequences and often contain only data from the snowy domain~\cite{pitropov2021cadc, boreas2023} or do not have enough labelled data in one of the domains to train an object detector~\cite{boreas2023, ithaca365, stf2020}. These datasets are generally inadequate and specifically fail to minimize domain discrepancies caused by confounding factors unrelated to snow.

An alternative way of creating paired datasets is via synthetic means. Real point clouds from snowy weather can be de-snowed~\cite{dror2018,ddior2022,4denoisenet2023,lilanet2018,lisnownet2022} to synthetically obtain the same scene under clear weather conditions. Conversely, snowfall simulation methods inject, drop, and attenuate points in real clear weather point clouds~\cite{noiseInjection2022,teufel2022,lisa2021,hahner2022}. With these approaches, only the desired weather conditions are introduced and everything else in the scene is preserved. Object annotations remain valid in the synthetic data, eliminating any additional labelling efforts that would otherwise be needed. However, de-snowing methods often falsely identify some points as snow and also fail to correctly identify all snow points. Moreover, they cannot easily reconstruct what the snow obstructs or capture the effects of attenuation. Snowfall simulation often employs numerical approximations to simplify the modelling of snow's physical properties, the LiDAR's optical system, and the laser beam's propagation through the scattering medium. Both de-snowing and snow synthesis currently do not address the accumulated snow on surfaces. Hence, generating synthetic data introduces an additional domain gap that cannot be distinguished from the weather domain gap we wish to isolate. 

These limitations have motivated us to extend the Canadian Adverse Driving Conditions (CADC)~\cite{pitropov2021cadc} with an equivalent amount of real data collected in clear weather from the same roads as CADC. This is feasible because we have access to a large amount of unused clear weather data collected in the same period as CADC, using the same vehicle platform. 
We call our extension CADC-clear and refer to the complete paired dataset with both weather domains as CADC+. Unlike other datasets that typically feature summer scenes, CADC-clear is recorded in winter, contemporaneously with CADC, with some sequences collected just one day apart from some of the CADC recordings. Although scenes inevitably differ due to the uncontrollable nature of the public environment, the temporal proximity helps minimize domain gaps caused by variations in roads and traffic agents, enabling the creation of a better-paired weather domain adaptation dataset.

To construct CADC-clear, we devise a matching method that minimizes temporal and spatial differences between two sequences. We aim for CADC-clear to have similar object, point, and motion distributions to CADC, minimizing domain shifts due to factors that are unrelated to snow. CADC-clear is thus a similar-sized clear weather extension to CADC, where each snowy sequence is paired with a clear weather sequence. An example pair of corresponding data frames is illustrated in \cref{fig:cadc_and_clear_examples}.

CADC contains 3D bounding box annotations for all 7000 LiDAR frames in 74 sequences. This is sufficient data to train a state-of-the-art (SOTA) 3D object detector like VoxelNeXt~\cite{chen2023voxelnext}. We lacked the budget to label all LiDAR frames in CADC-clear, but our success with sparse labelling~\cite{tang2025} allows us to use a more efficient approach. To ensure accurate evaluation, we fully label \emph{all} CADC-clear validation sequences and adopt a sparse labelling strategy only for the training sequences. For the latter, we label frames spaced at a fixed interval in \emph{all} sequences rather than labelling a subset of sequences. The intermediate frames remain unlabelled. A SOTA object detector can then be trained in semi-supervised fashion on all CADC-clear frames, using a mixture of ground truth and pseudo-labels produced by a model trained on the available ground-truth. We have previously shown that such a model can have performance close to that of one trained on 100\% human-annotated data~\cite{tang2025}. We further support this with the results shown in~\Cref{tab:sparse_labelling}.

In this work, we also present preliminary results of using CADC+. To evaluate the effect of snow on 3D object detection, we compare several models trained using different proportions of data from CADC and CADC-clear. We furthermore assess models trained on synthetic clear weather data generated by statistical~\cite{dror2018} and learning-based~\cite{lisnownet2022} de-snowing methods. Snowfall simulation methods are beyond the scope of these preliminary evaluations, due to not being directly compatible with our LiDAR sensor and requiring meticulous distribution fitting to yield meaningful results.

Our main contributions can be summarized as follows:

\begin{itemize}
\item We create CADC+, a paired-weather domain adaptation dataset extension of the existing CADC dataset. CADC+ is the first multi-modal dataset with 3D bounding box annotations for 74 sequence pairs from clear and snowy weather conditions.
\item We devise sequence matching and endpoint selection algorithms to minimize spatial discrepancies between paired trajectories.
\item We analyze the effect of snow on 3D object detection and evaluate the effectiveness of de-snowing point clouds to simulate clear weather conditions.
\end{itemize}

\section{Related Work}

\subsection{Snowy Weather Driving Datasets}

There has been a growing effort to build LiDAR datasets to improve perception in snowy weather~\cite{pitropov2021cadc, grounded2021, boreas2023, wads2022, ithaca365, stf2020}. The GROUNDED~\cite{grounded2021} and Boreas~\cite{boreas2023} datasets contain snowy data, but lack 3D bounding box annotations to support 3D object detection tasks. WADS~\cite{wads2022} offers pointwise semantic labels that could be converted into 3D bounding boxes, but it contains exclusively snowy scenes. CADC~\cite{pitropov2021cadc} provides 3D bounding box annotations and also features only snowy scenes. Ithaca365~\cite{ithaca365} and DENSE~\cite{stf2020} label both clear and snowy weather data with 3D bounding boxes. Ithaca365 only has approximately 4.7\,k labelled frames in clear weather and 1.8\,k in snow. DENSE offers slightly more, with about 6.2\,k clear weather frames and 4.7\,k snowy frames. However, the number of labelled frames in each weather domain remains smaller than that of the KITTI dataset~\cite{Geiger2013}, which is considered the minimum for training modern deep learning-based object detectors. 
While we have shown how unlabelled data may be utilized with sparse labelling and semi-supervised learning~\cite{tang2025}, sequences in Ithaca365 and DENSE do not follow the sparse labelling paradigm.
Additionally, DENSE's snowy scenes feature lighter snowfall than the heavier snowfall captured in the CADC and WADS datasets. Without sufficient labelled data in both clear and snowy weather domains, the impact of snowfall on 3D object detection performance cannot be adequately discerned. Moreover, none of these datasets attempt to minimize irrelevant domain differences by pairing sequences from the two domains, as we do in the present work.


\subsection{Synthetic Weather Datasets}

Synthetic clear weather data can be generated by de-snowing real snowy data. Due to the effects of backscattering, many approaches~\cite{dror2018, dsor2021, fcsor2018, lior2020, ddior2022, 4denoisenet2023, spatioTemporal2022, multiframeDesnowing2024, spatioTemporal2023} treat spurious snowflake returns as noise and develop de-noising methods to recover clear weather data from snowy recordings. Early de-noising approaches mainly rely on statistical filtering, using either the distance values of LiDAR points~\cite{dror2018, dsor2021, fcsor2018}, the intensity values~\cite{lior2020}, or a combination of both~\cite{ddior2022} to remove snowy outliers. With the release of the WADS dataset~\cite{wads2022}, which provides pointwise semantic labels for training and validation, there have emerged both supervised~\cite{weathernet2020, lilanet2018} and unsupervised~\cite{lisnownet2022, cycleganDesnowing2022} learning-based methods to segment out snowy points. More recently, researchers have used both spatial and temporal features in adjacent point clouds~\cite{4denoisenet2023, spatioTemporal2022, multiframeDesnowing2024, spatioTemporal2023}. While identifying and removing noisy snow points from a point cloud can effectively reduce backscattering effects, attenuation effects remain unaddressed. Additionally, imperfect de-noising algorithms will mistakenly remove points that are not snow, while also failing to remove some points that are snow. 

Synthetic snowy data can be generated by injecting, dropping, and attenuating points in real clear weather point clouds. One simplistic approach inserts uniform noise and discards object points that are obstructed~\cite{noiseInjection2022}. This is unrealistic due to the likelihood of particles producing returns being dependent on distance. Others have developed more sophisticated physics-based methods that model the distributions of snowflake diameter, density, and falling rate~\cite{teufel2022,lisa2021,hahner2022}. These distributions, which may be tuned to match real snowy datasets~\cite{teufel2022}, are used to generate snowflakes that are injected into the scene. 
Each LiDAR beam is then raytraced through the scattering medium to determine whether the original point should be dropped, attenuated, or replaced by a closer snow return. Despite the above efforts to simulate airborne snowflakes and also wet ground~\cite{hahner2022}, accumulated snow on roads and objects has yet to be modelled. 




\section{Method} 

\subsection{The Clear Weather Data}

\begin{figure}[t]
    \centering
    \includegraphics[width=0.65\linewidth]{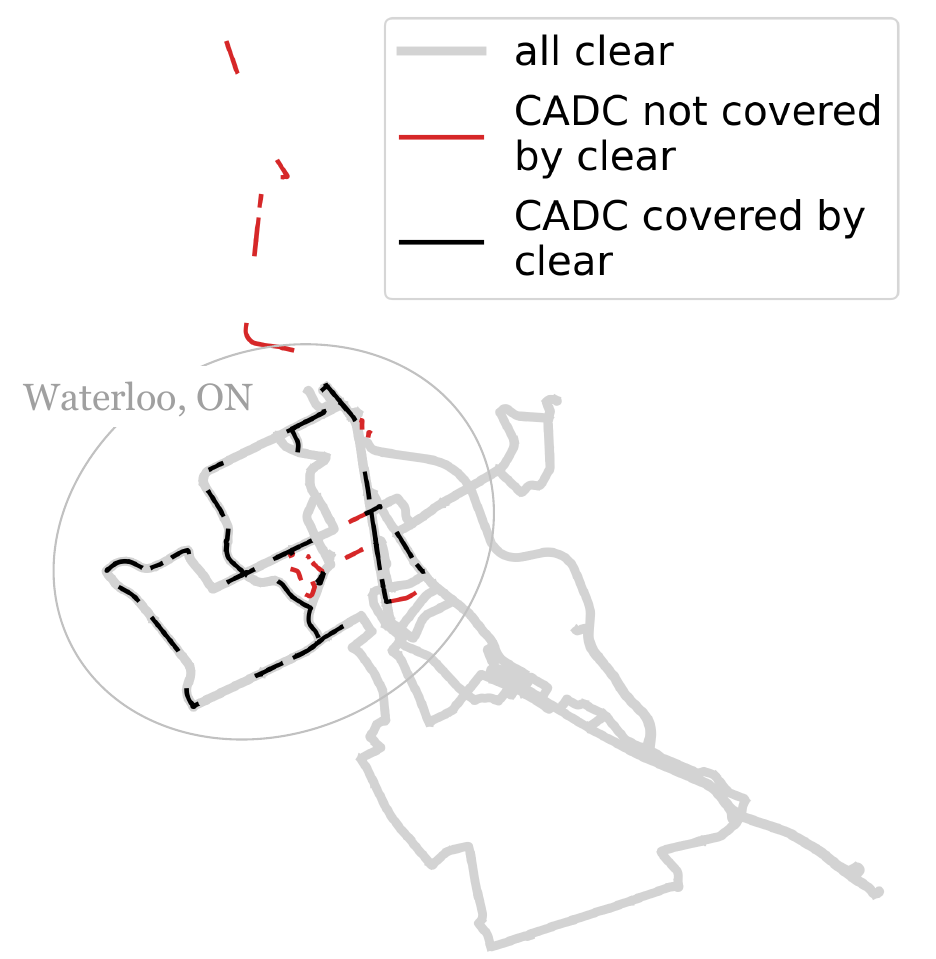}
    \caption{Coverage of CADC by clear sequences recorded in the same period.}
    \label{fig:clear_coverage}
    
\end{figure}

CADC comprises 74 labelled snowy weather sequences recorded in and around Waterloo, Ontario, on three days in 2018 and 2019, using our Autonomoose platform.\footnote{https://uwaterloo.ca/waterloo-intelligent-systems-engineering-lab/autonomoose} CADC-clear is drawn from more than 400 clear weather sequences recorded in and around Waterloo using the same platform, on four days in 2018 that interleave CADC's sequences.
\Cref{fig:clear_coverage} illustrates the roads covered by CADC and the clear sequences (grey lines). Black lines denote CADC sequences that have some coverage by a closely matching clear sequence, as defined in \cref{sec:pairing_methods}. Red lines denote CADC sequences for which there is no closely matching clear sequence. To match these CADC sequences, a clear sequence with a similar driving environment and road participants is chosen manually. 


\subsection{Matching Clear Sequences to CADC}
\label{sec:pairing_methods}

\begin{figure*}[t]
    \centering
    \begin{subfigure}[b]{0.33\linewidth}
        \centering
        \includegraphics[width=\linewidth]{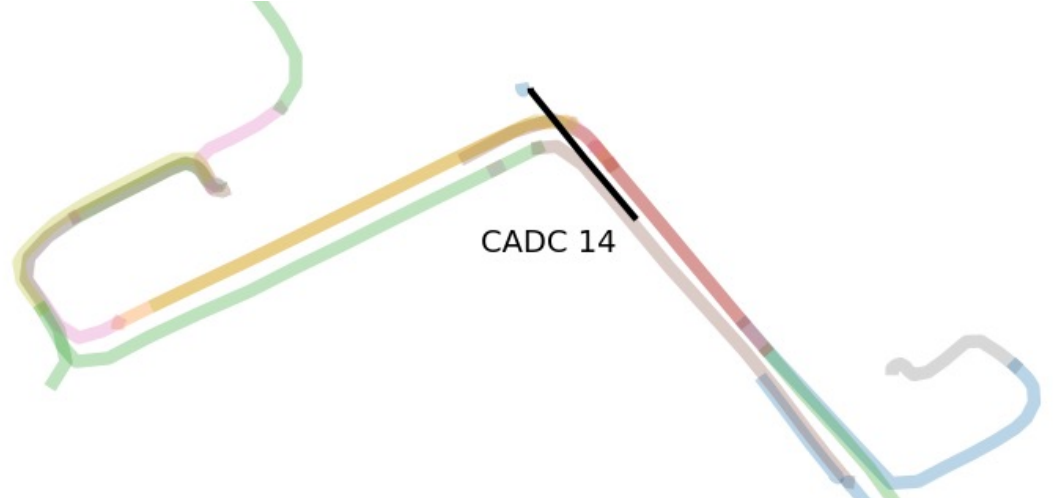}
        \caption{}
        \label{fig:matching_example_overall}
    \end{subfigure}
    \hfill
    \begin{subfigure}[b]{0.15\linewidth}
        \centering
        \includegraphics[width=\linewidth]{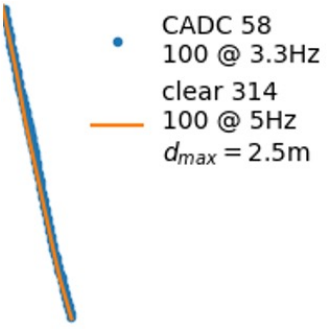}
        \caption{}
        \label{fig:matching_example_2}
    \end{subfigure}
    \hfill
    \begin{subfigure}[b]{0.15\linewidth}
        \centering
        \includegraphics[width=\linewidth]{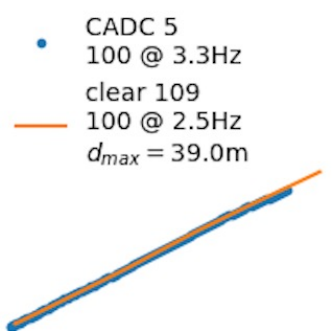}
        \caption{}
        \label{fig:matching_example_0}
    \end{subfigure}
    \hfill
    \begin{subfigure}[b]{0.15\linewidth}
        \centering
        \includegraphics[width=\linewidth]{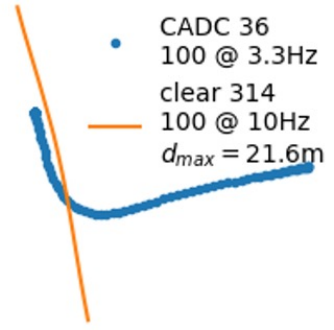}
        \caption{}
        \label{fig:matching_example_1}
    \end{subfigure}
    \hfill
    \begin{subfigure}[b]{0.15\linewidth}
        \centering
        \includegraphics[width=\linewidth]{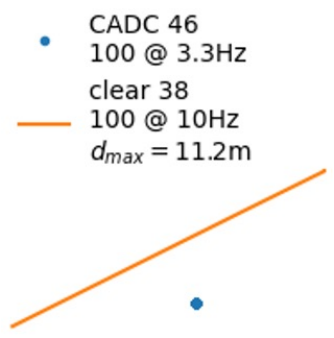}
        \caption{}
        \label{fig:matching_example_3}
    \end{subfigure}
    
    \caption{(a) Candidate clear sequences (various colours) near a CADC sequence (black). (b)--(e) Matched CADC (blue) and CADC-clear (orange) sequences. Each sequence is assigned an ID, beneath which we indicate the number of frames and sampling frequency after matching. $d_{\max}$ denotes the greatest minimum distance between two points in the paired sequences.}
    \label{fig:matches}
    
\end{figure*}


Ideally, we would like each snowy frame to have a matching clear frame that is observed from the same position and orientation, at the same time of day and with the same road users. The difference would then only be the snow. However, since the clear sequences were not recorded for the purpose of matching to snowy sequences, none of these requirements can be fully met. \Cref{fig:matching_example_overall} illustrates a typical situation: several clear weather sequences recorded in the vicinity of a CADC sequence, but none exactly align. Our matching therefore entails a definition of non-zero closeness that has multiple objectives.

For every snowy sequence, we try to find a clear sequence that can achieve the following informal objectives, where $\succ$ denotes “better than” and $\ggcurly$ denotes “much better than”:
 \begin{itemize}
     \item On the same road $\ggcurly$ on an adjacent road\label{obj:same-road}
     \item In the same direction $\succ$ in the opposite direction\label{obj:same-direction}
     \item In the same lane $\succ$ in an adjacent lane\label{it:same-lane}
     \item Ends near to ends of snowy sequence\label{obj:same-ends}
     \item Number of frames $\geq$ number of snowy frames\label{obj:same-frames}
     \item Equal or greater number of other road users than in snowy sequence\label{obj:same-road-users}
 \end{itemize}

We value longitudinal coverage over lateral displacement, so scalarizations based on simple distance metrics are not adequate. We therefore adopt the following filtering technique.

All the data sequences were collected at a constant sampling frequency, but the vehicle had different speeds and accelerations in each data collection run. The data is therefore sampled uniformly in time, but not in space. To make the necessary spatial comparison of snowy and clear sequences, we first construct discrete spatial models by interpolating constant distance line segments along all the recorded trajectories. The details are given in~\cref{sec:interpolation}.
%
%
We then calculate the longitudinal overlap of every clear and snowy spatial sequence pair using a function $\mathrm{cover}(s,c,\theta)$ that returns the fraction of points in the snowy sequence $s$ that are within a specified distance $\theta$ of a point in the clear sequence $c$, for a finite set of such distances $\Theta$. 
These correspond to the possible lateral distances between paths on the same road.
Function $\mathrm{cover}$ is defined in~\cref{sec:coverage}.

The set of clear sequences with greatest coverage of snowy sequence $s$, given lateral distance threshold $\theta$, is given by
\begin{equation}
    c^*_{s,\theta}=\argmax_{c\in\mathcal{C}}\mathrm{cover}(s,c,\theta).\label{eq:best-clear}
\end{equation}
The set of clear sequences with greatest nonzero coverage of snowy sequence $s$, for all $\theta\in\Theta$, is thus given by
\begin{equation}
    C^*_s=\bigcup\{c\in c^*_{s,\theta}\mid\mathrm{cover}(s,c,\theta)>0,\theta\in\Theta\}.
\end{equation}
Finally, we define a consistency function $\mathrm{cons}(s,c)$ that gives the subset of $\Theta$ for which $c$ is the clear sequence that has greatest nonzero coverage of snowy sequence $s$. Function $\mathrm{cons}$ is defined by
\begin{equation}
    \mathrm{cons}(s,c)=\{\theta\in\Theta\mid c\in c^*_{s,\theta},\mathrm{cover}(s,c,\theta)>0\}.
\end{equation}

Using the above notions, $C^*_s$ is refined into the following subsets, in decreasing order of desirability. A clear sequence is then chosen from the first nonempty subset:
\begin{enumerate}
    \item $\{c\in C^*_s\mid\mathrm{cover}(s,c,\theta)=1,\theta=\min(\Theta)\}$\label{set:perfect}
    \item $\{c\in C^*_s\mid\exists\theta\in\Theta\bullet\mathrm{cover}(s,c,\theta)\geq0.95\}$\label{set:good}
    \item $\{c\in C^*_s\mid|C^*_s|=1\}$\label{set:unique}
    \item $\{c\in C^*_s\mid|\mathrm{cons}(s,c)|>1\}$\label{set:consistent}
    \item $C^*_s$\label{set:nonzero}
\end{enumerate}

Informally, set~\ref{set:perfect} contains any clear sequences that are in the same lane and that completely cover the snowy sequence; set~\ref{set:good} contains any clear sequences that are in the same road and cover most of the snowy sequence. If there is just one clear sequence that has nonzero coverage of the snowy sequence, considering all lateral distances, it is in set~\ref{set:unique}. Set~\ref{set:consistent} contains clear sequences that have greatest coverage of the snowy sequence for more than one lateral distance.

When the first nonempty subset is not a singleton, a clear sequence is chosen by human inspection, based on its coverage of the clear sequence and the number and diversity of the other road users it contains. In the few cases that a snowy sequence has no matching sequence by the above process, i.e., $C^*_s=\varnothing$, a clear sequence is chosen that is as qualitatively similar to the snowy sequence as possible.

Having identified the clear sequences using spatial models, we must choose the actual data frames to use. CADC data sequences mostly contain 100 labelled frames sampled at 3.3\,Hz, with a handful of sequences containing fewer frames.
To match CADC and ensure that CADC-clear works well with our sequence-based auto-labelling techniques~\cite{tang2025}, we desire that all clear sequences have at least 100 frames. Given the inherently different driving speeds and accelerations that exist in clear and snowy weather, it is not feasible to simultaneously match the number of frames, the ends of the sequences, and the frame rates of the sequences. We therefore adopt the following procedure.

For each clear temporal sequence $c$ matched to a snowy temporal sequence $s$, we first find the set of frames in $c$ that are spatially closest to frames in $s$. From this set, we identify those with the earliest and latest timestamps, which demark a clear subsequence $c'$. The ends of $c'$ usually correspond to the ends of $s$, but not always, due to the vehicle remaining stationary in some sequences, and the existence of measurement errors. We require a minimum of 100 frames from $c$, sampled at regular intervals, so we find the maximum sampling interval $\in\{1,2,3,4,5\}$ that samples $c'$ into $\geq100$ frames. In rare cases where this produces more than 150 sampled frames, we increment the sampling interval and use exactly 100 frames. The sampled frames form the final clear sequence, which may have been sampled at 10, 5, 3.3, 2.5, or 2\,Hz, depending on the sampling interval.

\Cref{fig:matching_example_2} shows a clear sequence that matches a CADC sequence longitudinally, but its sampling rate is 5\,Hz to compensate for an otherwise insufficient frame count. The clear sequence in~\cref{fig:matching_example_0} has a sampling frequency of 2.5\,Hz to limit it to 100 frames. The clear sequence in~\cref{fig:matching_example_1}, however, only matches part of the CADC trajectory and has to extend beyond the closest matching frames to achieve 100 frames, thus increasing the maximum distance discrepancy. The clear sequence cannot be further shortened because its sampling frequency is already 10\,Hz, the maximum defined by our LiDAR's operating frequency. A similar situation arises when the data collection vehicle is stationary in CADC sequences, as illustrated in~\cref{fig:matching_example_3}.

\subsection{Dataset Labelling}
\label{sec:sparse_labelling}

\begin{table}[b]
\centering
\begin{tabular}{l|cc|cc@{}}
\toprule
\textbf{Model}                                                & \multicolumn{2}{l}{\textbf{L1 AP$_{\textsc{3D}}$\%}}                                          & \multicolumn{2}{l}{\textbf{L2 AP$_{\textsc{3D}}$\%}}                     \\ \midrule
Base$_{100\%}$                           & 70.15                                   &  $\Delta \%$                             & 48.29                        & $\Delta \%$         \\
Base$_{10\%}$ & 67.39                                   & {\color{red} -2.76} & 46.34                        & {\color{red} -1.96} \\
SSL$_{10\%}$             & 70.78                                   & {\color{grn} 0.63}  & 48.80                        & {\color{grn} 0.51}  \\ \bottomrule
\end{tabular}
\caption{Effectiveness of using semi-supervised learning with sparse labelling on CADC's Car class. We use the 3D IoU-based Level 1 (L1) and level 2 (L2) Average Precision (AP) metric from the Waymo Open Dataset~\cite{sun2020waymo}. The $\Delta \%$ column indicates the difference w.r.t. the Base$_{100\%}$ model.}
\label{tab:sparse_labelling}

\end{table}

We contracted a third-party labelling company to provide human-annotated 3D bounding boxes for our LiDAR point clouds. Since we only had budget to label approximately 30$\%$ of our data, we decided to fully label the validation split and label only 10\% of the training split. A fully labelled validation set is crucial for accurate evaluation, while our previous work has shown that it is possible to use semi-supervised learning (SSL) to achieve performance close to that of using 100\% human annotations with only 10\%~\cite{tang2025}. Importantly, instead of simply reducing the number of labelled sequences, which loses data diversity, we label the first and every subsequent tenth frame in all training sequences. 

To confirm that the SSL approach of~\cite{tang2025}, whose results are based on the Waymo Open Dataset~\cite{sun2020waymo}, will also work with CADC-clear, we apply it to the closely-related CADC dataset. This allows us to make comparisons with a model trained on 100\% human annotations. Using the VoxelNext~\cite{chen2023voxelnext} architecture and five-frame aggregation, we train a base model with all available annotations as our reference. We denote this model Base$_{100\%}$. We also train a base model using CADC sequences whose labels are subsampled to match those of CADC-clear. We denote this model Base$_{10\%}$. Finally, we generate pseudo-labels for the unlabelled frames using Base$_{10\%}$ and train a further model using both the pseudo-labels and the 10\% of human annotations. We denote this model SSL$_{10\%}$.

The detection results in~\Cref{tab:sparse_labelling} confirm the validity of our sparse labelling strategy. While the performance of Base$_{10\%}$ is worse than that of Base$_{100\%}$, as expected, the performance of SSL$_{10\%}$ slightly exceeds the performance of Base$_{100\%}$.

\subsection{Dataset Statistics}

\begin{figure}[t]
    \centering
    \begin{subfigure}[b]{0.4\linewidth}
        \centering
        \includegraphics[width=\linewidth]{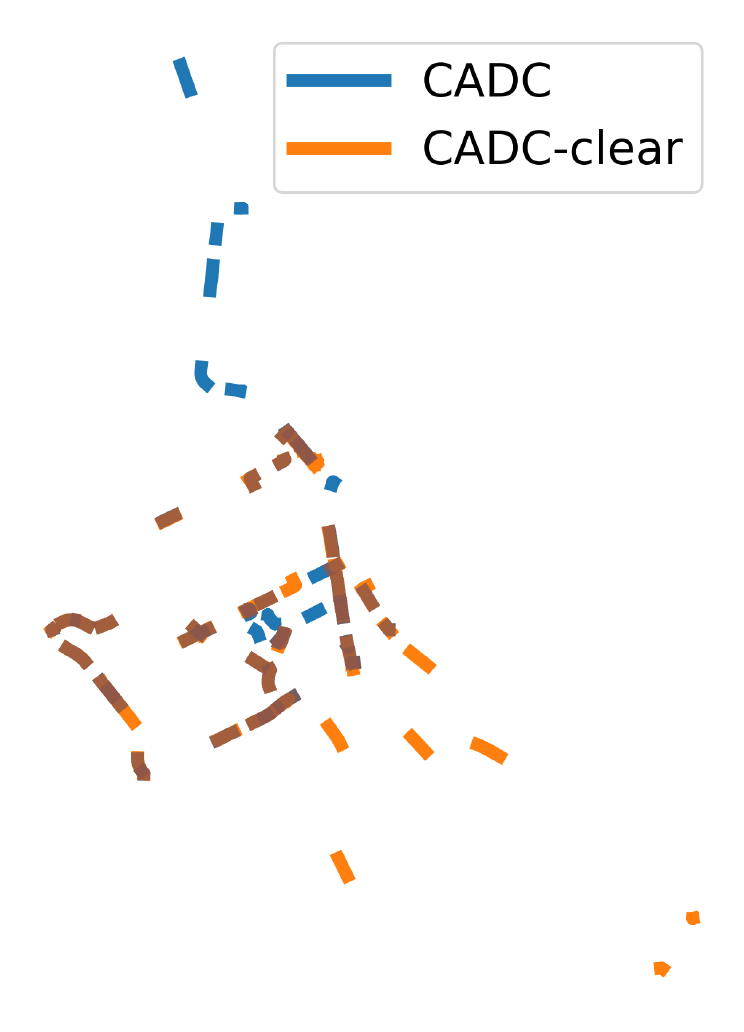}
        \caption{Training sets}
        \label{fig:matched_coverage_train}
    \end{subfigure}
    \hfill
    \begin{subfigure}[b]{0.4\linewidth}
        \centering
        \includegraphics[width=\linewidth]{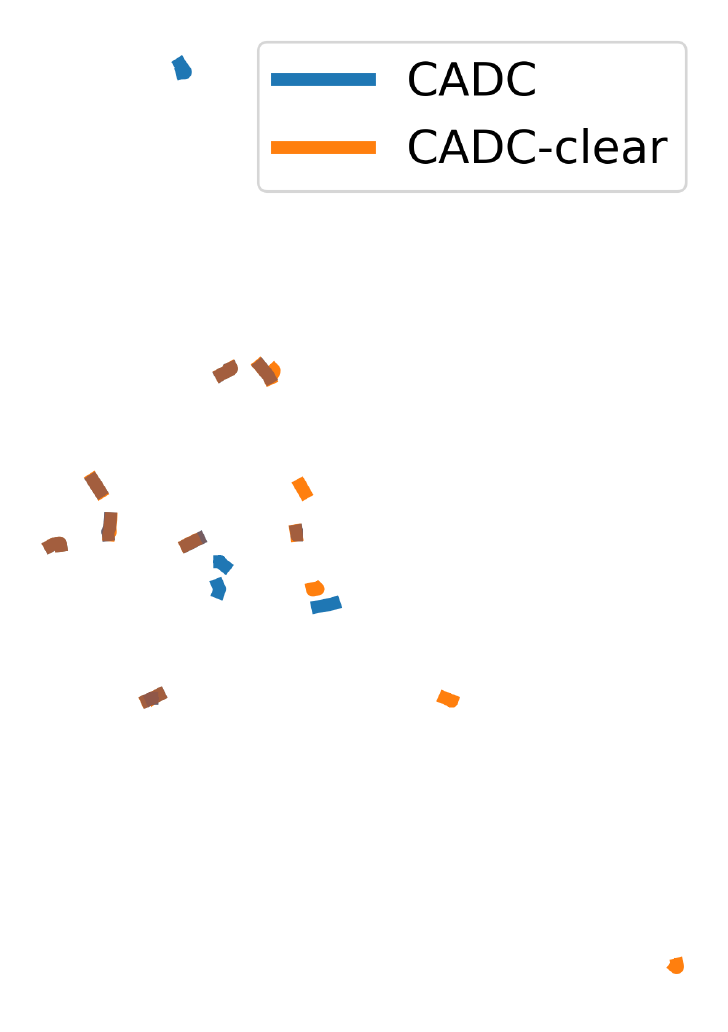}
        \caption{Validation sets}
        \label{fig:matched_coverage_val}
    \end{subfigure}
    
    \caption{Coverage of CADC and CADC-clear pairs after matching. Overlapping sequences are indicated in brown.}
    \label{fig:matched_coverage}
    
\end{figure}

\begin{figure}[b]
    \centering
    \includegraphics[width=\linewidth]{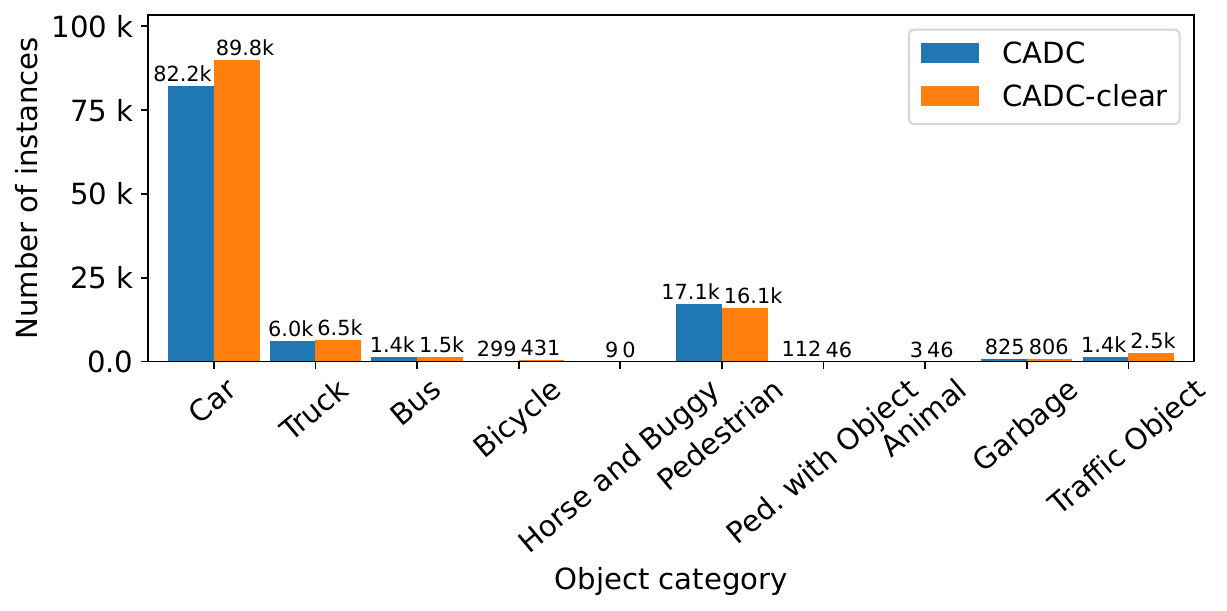}
    \caption{Object count by category in CADC and CADC-clear.}
    \label{fig:stats_obj_category}
    
\end{figure}

With one exception, all sequences in CADC are matched to a CADC-clear sequence of at least 100 frames.
The other case is where a single CADC sequence is matched to two half-length sequences from CADC-clear, for better alignment and data diversity.
CADC-clear thus has 60 training sequences and 15 validation sequences, where two of the validation sequences are half-length. \Cref{fig:matched_coverage} shows the resulting road coverage of CADC and CADC-clear, including both training and validation sets.

We were unable to closely match 21 of the 74 CADC sequences due to the lack of coverage illustrated in~\cref{fig:clear_coverage}. For 15 of these, we manually found a match from a different location with a similar environment. E.g., we matched a parking lot with a different parking lot. When there was no candidate sequence of similar traffic density, we favoured one with a higher density, in the hope that it would increase the amount and diversity of the data. For the remaining six unmatched sequences, we selected clear sequences matching the nature of the CADC road agents. E.g., a pedestrian-heavy intersection on the university campus was matched to a pedestrian-heavy road intersection in the city.

\begin{figure}[t]
    \centering

    \begin{subfigure}[t]{0.46\linewidth}
        \centering
        \includegraphics[width=\linewidth]{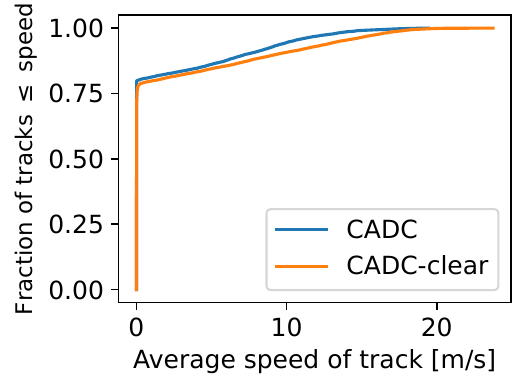}
        \caption{}
        \label{fig:stats_motion_1_car}
    \end{subfigure}
    \hfill
    \begin{subfigure}[t]{0.46\linewidth}
        \centering
        \includegraphics[width=\linewidth]{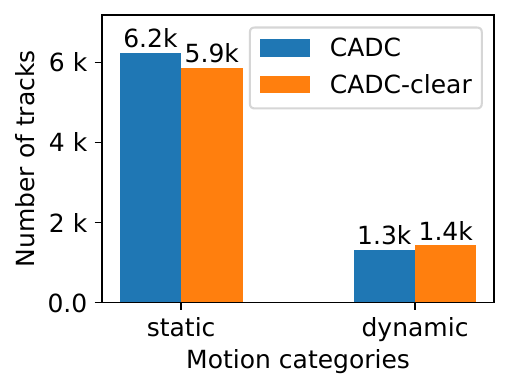}
        \caption{}
        \label{fig:stats_motion_2_car}
    \end{subfigure}

    \begin{subfigure}[t]{0.46\linewidth}
        \centering
        \includegraphics[width=\linewidth]{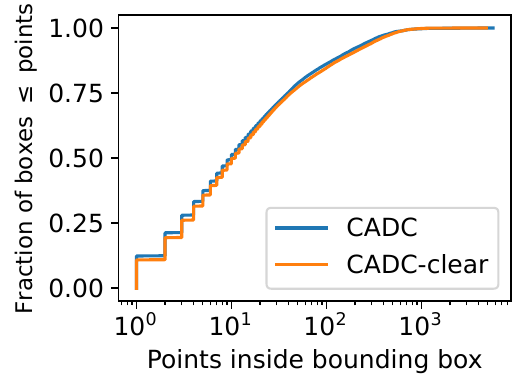}
        \caption{}
        \label{fig:stats_cdfs_1_car}
    \end{subfigure}
    \hfill
    \begin{subfigure}[t]{0.46\linewidth}
        \centering
        \includegraphics[width=\linewidth]{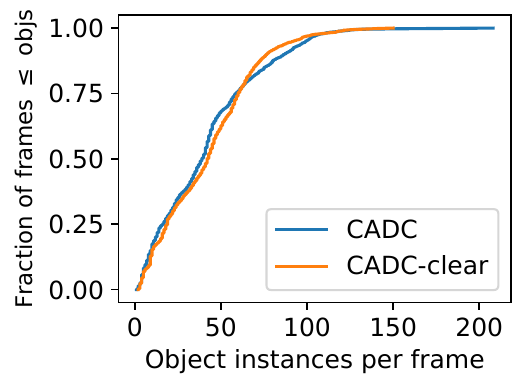}
        \caption{}
        \label{fig:stats_cdfs_2_car}
    \end{subfigure}
    
    \caption{Distributions of point count per annotation, object count per frame average speed, and motion for cars in CADC and CADC-clear.}
    \label{fig:stats}
\end{figure}

To compare the clear and snowy datasets, we first subsample CADC sequences to match the sparsity of CADC-clear. We thus obtain the distributions shown in~\cref{fig:stats_obj_category,fig:stats}. \Cref{fig:stats_obj_category} shows that the object instance count in CADC-clear is slightly higher than in CADC, which aligns with our expectation of increased road user presence in clear weather. The slight lack of pedestrian instances can be attributed to the six CADC sequences for which we could not find a close match in clear weather. These sequences were recorded on the university campus, where the presence of students heading to class results in a higher density of pedestrians compared to city streets. 

In~\cref{fig:stats_motion_1_car}, we plot the empirical cumulative distributions of the average estimated speed of cars along tracks created by gathering all bounding boxes with the same ID. We see that the CADC-clear distribution is shifted to the right, in part due to having fewer stationary objects, as indicated by the vertical parts of the distributions near zero speed, but also because vehicles tend to travel faster in clear weather. Using a threshold of $0.2$\,m/s to distinguish stationary and dynamic vehicles, as defined in the Waymo Open Dataset evaluation~\cite{sun2020waymo}, we confirm in~\cref{fig:stats_motion_2_car} that the missing cars in CADC-clear are mostly stationary, likely coming from the unmatched parking lots.

Although CADC-clear has a slightly higher total number of objects, its distributions of point count per bounding box (\cref{fig:stats_cdfs_1_car}) and object count per frame (\cref{fig:stats_cdfs_2_car}) are almost indistinguishable from those of CADC. This alignment is important for CADC+ to serve as a weather domain adaptation dataset and enable objective evaluation of snow's impact on 3D object detection with minimal bias from confounding factors unrelated to snow.

\section{Experiments}




Multiple prior works~\cite{Rasshofer2011, Jokela2019, Kutila2020, surveyZhang2023} have demonstrated that snowy conditions affect LiDAR point clouds, but \emph{how hard is snow} for LiDAR-based 3D object detection? In this section, we provide some preliminary answers using CADC+. Specifically, we compare the clear and snowy weather detection performance of multiple models trained on varying proportions of snowy and clear data from CADC and CADC-clear, respectively. By cross-evaluating the models in this way, we establish insightful trends of how snow affects 3D object detection performance. To give additional context and motivate the need for CADC+, we also present the results of cross-evaluation experiments using synthetic clear data provided by DROR~\cite{dror2018} and LiSnowNet~\cite{lisnownet2022}. DROR is a common statistical baseline approach, while LiSnowNet is a SOTA learning-based approach. 

To ensure correct and fair results, we downsample the labels of CADC's training split to match the sparse labelling of the CADC-clear training split.
The validation splits of both CADC and CADC-clear remain fully labelled. To create training sets with different proportions of snowy and clear data, we first subsample the downsampled CADC and sparsely labelled CADC-clear training data into separate splits with fewer labels. We create a 0.25 split by selecting the first frame and every fourth labelled frame in all sequences. We create a 0.5 split by selecting the first frame and every second labelled frame in all sequences. We create a 0.75 split by discarding every fourth labelled frame in all sequences. In addition to the pure snowy and clear training datasets (1.0 splits), we construct mixed training datasets by combining the CADC 0.25 split with the CADC-clear 0.75 split, the CADC 0.5 split with the CADC-clear 0.5 split, and the CADC 0.75 split with the CADC-clear 0.25 split. All mixed training datasets therefore have approximately the same number of labels as the 1.0 splits, which label every tenth frame in all sequences.

For every combination of human-annotated clear and snowy training data, we train a Base model using only the human-annotated labels and an SSL model using a combination of the same human-annotated labels plus high-quality pseudo-labels produced by the corresponding Base model applied to the unlabelled frames. All models are trained on the dominant Car class, using the 5-frame aggregated VoxelNeXt architecture~\cite{chen2023voxelnext}. To minimize confounding factors, we normalize the number of training epochs with respect to the amount of data in each experiment. Base models are thus trained for approximately 400 epochs and SSL models are trained for approximately 40 epochs, with small variations due to the slightly different numbers of snowy and clear labels in each experiment. We evaluate all models on snowy weather (CADC's validation set) and clear weather (CADC-clear's validation set) using Waymo's L1 and L2 3D IoU-based AP metric~\cite{sun2020waymo}. 

For our de-snowing experiments, we construct synthetic clear data by applying DROR and LiSnowNet to the downsampled CADC training data.
The resulting datasets are denoted de-snowed$_{\text{DROR}}$ and de-snowed$_{\text{LiSnowNet}}$, respectively, which we use to train Base and SSL models in the way described above. We compare these models to models trained using CADC+ in~\Cref{tab:desnowing}.

\subsection{The Impact of Snow on 3D Object Detection}

\begin{figure}[t]
    \centering
    \includegraphics[width=\linewidth]{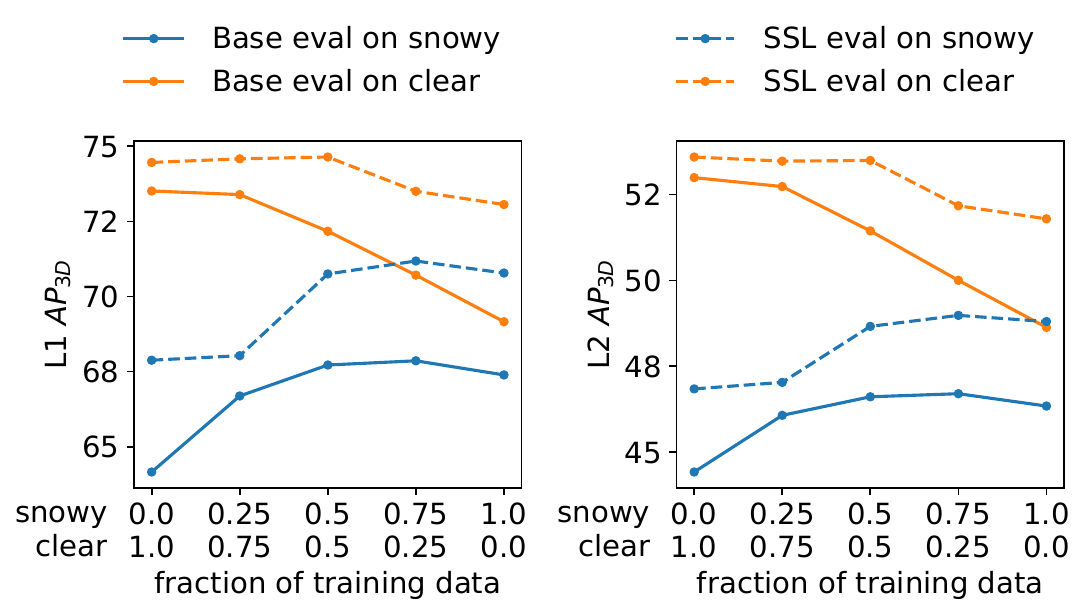}
    \caption{Cross-evaluation of Base and SSL models trained on various mixtures of snowy and clear data.
    }
    \label{fig:results}
    
\end{figure}

\begin{table*}[h]
\begin{minipage}[c]{0.67\textwidth}
\centering
\centering
\begin{tabular}{@{}c|cc|cc|cc|cc@{}}
\toprule
 \multicolumn{1}{c}{\textbf{Base model}} & \multicolumn{4}{c}{\textbf{Eval on snowy}} & \multicolumn{4}{c}{\textbf{Eval on clear}} \\
training data & \multicolumn{2}{l}{\textbf{L1 AP$_{\textsc{3D}}$\%}} & \multicolumn{2}{l|}{\textbf{L2 AP$_{\textsc{3D}}$\%}} & \multicolumn{2}{l}{\textbf{L1 AP$_{\textsc{3D}}$\%}} & \multicolumn{2}{l}{\textbf{L2 AP$_{\textsc{3D}}$\%}} \\ \midrule
clear & 64.16 & $\Delta\%$ & 44.42 & $\Delta\%$ & 73.51 & $\Delta\%$ & 52.99 & $\Delta\%$ \\
de-snowed$_{\text{DROR}}$ & 62.98 & {\color{red} -1.19} & 42.93 & {\color{red} -1.49} & 66.80 & {\color{red} -6.72} & 46.26 & {\color{red} -6.73} \\
de-snowed$_{\text{LiSnowNet}}$ & 64.88 & {\color{grn} 0.72} & 44.44 & {\color{grn} 0.02} & 68.82 & {\color{red} -4.69} & 48.13 & {\color{red} -4.86} \\
snowy & 67.39 & {\color{grn} 3.23} & 46.34 & {\color{grn} 1.92} & 69.16 & {\color{red} -4.35} & 48.63 & {\color{red} -4.36} \\ \bottomrule
\end{tabular}

\vspace{1em}
\centering
\begin{tabular}{@{}c|cc|cc|cc|cc@{}}
\toprule
 \multicolumn{1}{c}{\textbf{SSL model}} & \multicolumn{4}{c}{\textbf{Eval on snowy}} & \multicolumn{4}{c}{\textbf{Eval on clear}} \\
training data & \multicolumn{2}{l}{\textbf{L1 AP$_{\textsc{3D}}$\%}} & \multicolumn{2}{l|}{\textbf{L2 AP$_{\textsc{3D}}$\%}} & \multicolumn{2}{l}{\textbf{L1 AP$_{\textsc{3D}}$\%}} & \multicolumn{2}{l}{\textbf{L2 AP$_{\textsc{3D}}$\%}} \\ \midrule
clear & 67.88 & $\Delta\%$ & 46.84 & $\Delta\%$ & 74.46 & $\Delta\%$ & 53.59 & $\Delta\%$ \\
de-snowed$_{\text{DROR}}$ & 69.61 & {\color{grn} 1.73} & 47.44 & {\color{grn} 0.60} & 71.26 & {\color{red} -3.20} & 49.74 & {\color{red} -3.85} \\
de-snowed$_{\text{LiSnowNet}}$ & 70.43 & {\color{grn} 2.55} & 48.36 & {\color{grn} 1.51} & 72.74 & {\color{red} -1.72} & 51.33 & {\color{red} -2.27} \\
snowy & 70.78 & {\color{grn} 2.91} & 48.80 & {\color{grn} 1.96} & 73.06 & {\color{red} -1.40} & 51.79 & {\color{red} -1.80} \\ \bottomrule
\end{tabular}
\end{minipage}
\begin{minipage}[c]{0.3\textwidth}
\caption{Effectiveness of de-snowing on Base and SSL models. CADC data de-snowed by DROR and LiSnowNet is denoted de-snowed$_{\text{DROR}}$ and de-snowed$_{\text{LiSnowNet}}$, respectively. $\Delta \%$ indicates the difference w.r.t. the clear model.}
\label{tab:desnowing}
\end{minipage}
\end{table*}

\Cref{fig:results} presents the detection performance of Base (solid lines) and SSL (dashed lines) models trained with an increasing proportion of snowy data, evaluated on snowy (blue lines) and clear weather (orange lines). We note that the trends for L1 and L2 are consistent, with the more difficult L2 having lower performance, as expected. The following comments therefore apply to both L1 and L2.

The results in~\cref{fig:results} confirm our experience that semi-supervised learning (SSL, dashed lines) generally improves on the performance of a Base model (solid lines), due to the additional data diversity coming from unlabelled frames. We also see that performance on snowy weather (blue lines) is generally worse than on clear weather (orange lines) for the same model type, indicating that snowy objects are more difficult to detect. Given that our experiments control for factors unrelated to snow, including for overfitting and underfitting, this observation supports the conclusion that snow introduces aleatoric uncertainty.

The results in~\cref{fig:results} also support the conclusion that snow introduces epistemic uncertainty, i.e., that snowy objects constitute a distinct domain. The evidence for this is that the detection of snowy objects (blue lines) improves with an increasing proportion of snowy training data, up to about 75\%. The fact that training on 100\% snowy data does not maximize the detection of snowy objects is likely related to the presence of aleatoric uncertainty and the particular composition of the snowy validation data, which inevitably contains some objects that are not affected by snow. In contrast, the detection of clear objects does appear to be maximized by training on 100\% clear data, especially in the case of Base models. In the case of SSL models, the very slight peak in L1 AP seen with 50\% snowy training data is not present in L2, suggesting the peak is just the result of a fortuitous combination of data and learning. However, this model has near-maximum performance on both snowy and clear data, making it the best choice for a detector that encounters both domains.

\subsection{The Effectiveness of De-snowing for 3D Object Detection}

\Cref{tab:desnowing} shows the results of training Base and SSL models on de-snowed data produced by DROR and LiSnowNet, compared with models trained on real clear and snowy data from CADC+. The goal of these de-snowing techniques is to create realistic clear data, so we judge them by how closely the performance of their corresponding models resembles that of models trained on real clear data.

Considering first the Base models evaluated on snowy weather, we see that the de-snowed$_\text{LiSnowNet}$ model has performance close to that of the clear model, suggesting that the de-snowing is effective. The de-snowed$_\text{DROR}$ model is less close, in line with our understanding that DROR is less effective, but is still closer to the clear model than to the snowy model. However, the evaluation of the Base models on clear weather tells a different story. The performances of the de-snowed$_\text{LiSnowNet}$ and de-snowed$_\text{DROR}$ models are much closer to the performance of the snowy model than to the clear model, suggesting that the de-snowed data is not realistic and that the reduction of performances seen in the evaluation on snowy weather is simply the result of a general degradation in performance.

Looking now at the SSL models evaluated on snowy weather, we see that with the exception of the L2 performance of the de-snowed$_\text{DROR}$ model, the de-snowed models are closer to the snowy model than to the clear model. Evaluation of the SSL models on clear weather follows the same pattern seen with the Base models: de-snowed models are all significantly closer to the snowy model than to the clear model. The SSL results confirm that the chosen de-snowing techniques do not produce realistic clear data and that semi-supervised learning cannot compensate for the lack of realism, thus motivating the need for datasets like CADC+.

\section{Conclusion}

We have introduced the paired-weather domain adaptation dataset CADC+ and presented preliminary experiments that use it to evaluate the nature and effect of snow on 3D object detection. We have also used CADC+ to show that two popular de-snowing techniques do not produce adequately realistic clear data from snowy data.
Evaluating the effectiveness of physics-based snowfall simulation from clear weather data is beyond the scope of the present work, but would be a natural and interesting extension in the future. Other potential research directions include designing weather-invariant 3D object detectors, improving current de-snowing methods, and developing weather-based style transfer methods.

In terms of limitations, we note that CADC+'s clear weather set, CADC-clear, consists of sequences sampled at different frequencies and that while its validation split is fully labelled, its training split is labelled on every tenth frame. To mitigate these limitations, we provide unlabelled sensor data at 10\,Hz and have shown that using semi-supervised learning can obtain pseudo-labels of quality close to human annotations. We nevertheless recognize that these mitigations might restrict certain research use cases.

\section*{Acknowledgement}
We thank Micha{\l} Antkiewicz for contributions during the dataset labelling process and Adrian Chow for contributions in label validation. The dataset labelling was funded by Transport Canada’s 2024-2026 Enhanced Road Safety Transfer Payment Program (ERSTPP) as part of the project “Enhancing Driving Automation Safety in Canada: A Traffic Data-Driven Approach for Advanced Driver Assistance System and Connected and Autonomous Vehicle Test Scenarios” at the University of Waterloo.

\bibliographystyle{IEEEtranS}
{\footnotesize
\bibliography{main}}

\begin{thebibliography}{10}
\providecommand{\url}[1]{#1}
\csname url@samestyle\endcsname
\providecommand{\newblock}{\relax}
\providecommand{\bibinfo}[2]{#2}
\providecommand{\BIBentrySTDinterwordspacing}{\spaceskip=0pt\relax}
\providecommand{\BIBentryALTinterwordstretchfactor}{4}
\providecommand{\BIBentryALTinterwordspacing}{\spaceskip=\fontdimen2\font plus
\BIBentryALTinterwordstretchfactor\fontdimen3\font minus \fontdimen4\font\relax}
\providecommand{\BIBforeignlanguage}[2]{{%
\expandafter\ifx\csname l@#1\endcsname\relax
\typeout{** WARNING: IEEEtranS.bst: No hyphenation pattern has been}%
\typeout{** loaded for the language `#1'. Using the pattern for}%
\typeout{** the default language instead.}%
\else
\language=\csname l@#1\endcsname
\fi
#2}}
\providecommand{\BIBdecl}{\relax}
\BIBdecl

\bibitem{noiseInjection2022}
A.~D. Abu-Shaqra, N.~J. Abu-Alrub, and N.~A. Rawashdeh, ``Object detection in degraded {LiDAR} signals by synthetic snowfall noise for autonomous driving,'' in \emph{Autonomous Systems: Sensors, Processing and Security for Ground, Air, Sea and Space Vehicles and Infrastructure 2022}, vol. 12115, International Society for Optics and Photonics.\hskip 1em plus 0.5em minus 0.4em\relax SPIE, 2022, pp. 121\,150K:1--8.

\bibitem{fcsor2018}
H.~Balta, J.~Velagic \emph{et~al.}, ``Fast statistical outlier removal based method for large {3D} point clouds of outdoor environments,'' \emph{IFAC-PapersOnLine}, vol.~51, no.~22, pp. 348--353, 2018, 12th IFAC Symposium on Robot Control SYROCO 2018.

\bibitem{cycleganDesnowing2022}
J.~Bergius and J.~Holmblad, ``{LiDAR} point cloud de-noising for adverse weather,'' Master's thesis, , School of Information Technology, 2022.

\bibitem{stf2020}
M.~Bijelic, T.~Gruber \emph{et~al.}, ``Seeing through fog without seeing fog: Deep multimodal sensor fusion in unseen adverse weather,'' in \emph{Proceedings of the IEEE/CVF Conference on Computer Vision and Pattern Recognition}, 2020, pp. 11\,682--11\,692.

\bibitem{boreas2023}
K.~Burnett, D.~J. Yoon \emph{et~al.}, ``{Boreas}: A multi-season autonomous driving dataset,'' \emph{The International Journal of Robotics Research}, vol.~42, no. 1-2, pp. 33--42, 2023.

\bibitem{dror2018}
N.~Charron, S.~Phillips, and S.~L. Waslander, ``De-noising of lidar point clouds corrupted by snowfall,'' in \emph{2018 15th Conference on Computer and Robot Vision (CRV)}, 2018, pp. 254--261.

\bibitem{chen2023voxelnext}
Y.~Chen, J.~Liu \emph{et~al.}, ``{VoxelNeXt}: Fully sparse {VoxelNet} for {3D} object detection and tracking,'' in \emph{IEEE/CVF Conference on Computer Vision and Pattern Recognition (CVPR)}, 2023, pp. 21\,674--21\,683.

\bibitem{ithaca365}
C.~A. Diaz-Ruiz, Y.~Xia \emph{et~al.}, ``Ithaca365: Dataset and driving perception under repeated and challenging weather conditions,'' in \emph{IEEE/CVF Conference on Computer Vision and Pattern Recognition (CVPR)}, June 2022, pp. 21\,383--21\,392.

\bibitem{Geiger2013}
A.~Geiger, P.~Lenz, C.~Stiller, and R.~Urtasun, ``Vision meets robotics: The {KITTI} dataset,'' \emph{The International Journal of Robotics Research}, vol.~32, no.~11, pp. 1231--1237, 2013.

\bibitem{hahner2022}
M.~Hahner, C.~Sakaridis \emph{et~al.}, ``{LiDAR} snowfall simulation for robust {3D} object detection,'' in \emph{IEEE/CVF Conference on Computer Vision and Pattern Recognition (CVPR)}, June 2022, pp. 16\,364--16\,374.

\bibitem{weathernet2020}
R.~Heinzler, F.~Piewak, P.~Schindler, and W.~Stork, ``{CNN-Based} lidar point cloud de-noising in adverse weather,'' \emph{IEEE Robotics and Automation Letters}, vol.~5, no.~2, pp. 2514--2521, 2020.

\bibitem{Jokela2019}
M.~Jokela, M.~Kutila, and P.~Pyykönen, ``Testing and validation of automotive point-cloud sensors in adverse weather conditions,'' \emph{Applied Sciences}, vol.~9, no.~11, 2019.

\bibitem{lisa2021}
V.~Kilic, D.~Hegde \emph{et~al.}, ``Lidar light scattering augmentation ({LISA}): Physics-based simulation of adverse weather conditions for {3D} object detection,'' 2021, arXiv:\,2107.07004.

\bibitem{dsor2021}
A.~Kurup and J.~Bos, ``{DSOR}: A scalable statistical filter for removing falling snow from {LiDAR} point clouds in severe winter weather,'' 2021, arXiv:\,2109.07078.

\bibitem{wads2022}
------, ``Winter adverse driving dataset ({WADS}): year three,'' in \emph{Autonomous Systems: Sensors, Processing and Security for Ground, Air, Sea and Space Vehicles and Infrastructure 2022}, vol. 12115, International Society for Optics and Photonics.\hskip 1em plus 0.5em minus 0.4em\relax SPIE, 2022, pp. 031\,207:1--17.

\bibitem{Kutila2020}
M.~Kutila, P.~Pyykönen \emph{et~al.}, ``Benchmarking automotive {LiDAR} performance in arctic conditions,'' in \emph{IEEE International Conference on Intelligent Transportation Systems (ITSC)}, 2020, pp. 1--8.

\bibitem{spatioTemporal2022}
B.~Li, J.~Li \emph{et~al.}, ``De-snowing lidar point clouds with intensity and spatial-temporal features,'' in \emph{International Conference on Robotics and Automation (ICRA)}, 2022, pp. 2359--2365.

\bibitem{grounded2021}
T.~Ort, I.~Gilitschenski, and D.~Rus, ``{GROUNDED}: The localizing ground penetrating radar evaluation dataset.'' in \emph{Robotics: Science and Systems}, vol.~2, 2021.

\bibitem{lior2020}
J.-I. Park, J.~Park, and K.-S. Kim, ``Fast and accurate desnowing algorithm for {LiDAR} point clouds,'' \emph{IEEE Access}, vol.~8, pp. 160\,202--160\,212, 2020.

\bibitem{lilanet2018}
F.~Piewak, P.~Pinggera \emph{et~al.}, ``Boosting {LiDAR}-based semantic labeling by cross-modal training data generation,'' in \emph{European Conference on Computer Vision (ECCV)}, September 2018.

\bibitem{pitropov2021cadc}
M.~Pitropov, D.~E. Garcia \emph{et~al.}, ``Canadian adverse driving conditions dataset,'' \emph{The International Journal of Robotics Research}, vol.~40, no. 4-5, pp. 681--690, 2021.

\bibitem{Rasshofer2011}
R.~H. Rasshofer, M.~Spies, and H.~Spies, ``Influences of weather phenomena on automotive laser radar systems,'' \emph{Advances in Radio Science}, vol.~9, pp. 49--60, 2011.

\bibitem{4denoisenet2023}
A.~Seppänen, R.~Ojala, and K.~Tammi, ``{4DenoiseNet}: Adverse weather denoising from adjacent point clouds,'' \emph{IEEE Robotics and Automation Letters}, vol.~8, no.~1, pp. 456--463, 2023.

\bibitem{sun2020waymo}
P.~Sun, H.~Kretzschmar \emph{et~al.}, ``Scalability in perception for autonomous driving: {W}aymo {O}pen {D}ataset,'' in \emph{IEEE/CVF Conference on Computer Vision and Pattern Recognition (CVPR)}, 2020.

\bibitem{tang2025}
M.~Q. Tang, V.~Abdelzad \emph{et~al.}, ``{3D} object detection with track-based auto-labelling using very sparsely labelled data,'' in \emph{IEEE International Conference on Intelligent Transportation Systems (ITSC)}.\hskip 1em plus 0.5em minus 0.4em\relax IEEE, 2024, to appear.

\bibitem{teufel2022}
S.~Teufel, G.~Volk, A.~Von~Bernuth, and O.~Bringmann, ``Simulating realistic rain, snow, and fog variations for comprehensive performance characterization of {LiDAR} perception,'' in \emph{2022 IEEE 95th Vehicular Technology Conference: (VTC2022-Spring)}, 2022, pp. 1--7.

\bibitem{spatioTemporal2023}
W.~Wang, T.~Yang, Y.~Du, and Y.~Liu, ``Snow removal for {LiDAR} point clouds with spatio-temporal conditional random fields,'' \emph{IEEE Robotics and Automation Letters}, vol.~8, no.~10, pp. 6739--6746, 2023.

\bibitem{ddior2022}
W.~Wang, X.~You \emph{et~al.}, ``A scalable and accurate de-snowing algorithm for {LiDAR} point clouds in winter,'' \emph{Remote Sensing}, vol.~14, no.~6, 2022.

\bibitem{multiframeDesnowing2024}
X.~Yan, J.~Yang \emph{et~al.}, ``Denoising framework based on multiframe continuous point clouds for autonomous driving {LiDAR} in snowy weather,'' \emph{IEEE Sensors Journal}, vol.~24, no.~7, pp. 10\,515--10\,527, 2024.

\bibitem{lisnownet2022}
M.-Y. Yu, R.~Vasudevan, and M.~Johnson-Roberson, ``{LiSnowNet}: Real-time snow removal for {LiDAR} point clouds,'' in \emph{IEEE/RSJ International Conference on Intelligent Robots and Systems (IROS)}, 2022, pp. 6820--6826.

\bibitem{surveyZhang2023}
Y.~Zhang, A.~Carballo, H.~Yang, and K.~Takeda, ``Perception and sensing for autonomous vehicles under adverse weather conditions: A survey,'' \emph{ISPRS Journal of Photogrammetry and Remote Sensing}, vol. 196, pp. 146--177, 2023.

\end{thebibliography}

\appendix
\subsection{Interpolation}\label{sec:interpolation}
Let $w^\mathrm{t}_i$ denote the spatial (GPS) coordinates of frame $i$ in a collected data sequence $s$, where $t$ emphasizes that consecutive frames are separated by constant time $\delta^\mathrm{t}$. Then, let $w^\mathrm{d}_j$ denote the coordinates of point $j$ in a corresponding spatial model of $s$, where $\mathrm{d}$ emphasizes that consecutive points are separated by constant distance $\delta^\mathrm{d}$. Points $w^\mathrm{d}_j$ are constructed by interpolation from frames $w^\mathrm{t}_i$ as follows:
\begin{equation}
\begin{split}
    w^\mathrm{d}_0 &= w^\mathrm{t}_0\\
    w^\mathrm{d}_j &= \alpha(w^\mathrm{t}_i-w^\mathrm{t}_{i-1}) + w^\mathrm{t}_{i-1}\qquad i,j>0,
\end{split}\label{eq:dist-from-time}
\end{equation}
with $i,j$ constrained by
\begin{align}
0 &< j\delta^\mathrm{d}\qquad i=1\\
\sum_{k=1}^{i-1} \lVert w^\mathrm{t}_k-w^\mathrm{t}_{k-1}\rVert &< j\delta^\mathrm{d}\qquad i>1\\
\sum_{k=1}^{i} \lVert w^\mathrm{t}_k-w^\mathrm{t}_{k-1}\rVert &\geq j\delta^\mathrm{d},
\end{align}
and
\begin{equation}
    \alpha = \frac{j\delta^\mathrm{d}-\sum^{i-1}_{k=1}\lVert w^\mathrm{t}_k-w^\mathrm{t}_{k-1}\rVert}{\lVert w^\mathrm{t}_i-w^\mathrm{t}_{i-1}\rVert}.
\end{equation}
\subsection{Coverage}\label{sec:coverage}
Let $\mathcal{S}$ and $\mathcal{C}$ denote the sets of snowy and clear spatial sequences, respectively, and let $\Theta$ be the set of lateral distances we wish to consider. We thus define a coverage function $\mathrm{cover}:\mathcal{S}\times\mathcal{C}\times\Theta\to[0,1]$ that returns the fraction of points in a snowy sequence $s\in\mathcal{S}$ that are within $\theta\in\Theta$\,m of a point in clear sequence $c\in\mathcal{C}$. The function $\mathrm{cover}$ is implemented as:
\begin{equation}
\mathrm{cover}(s,c,\theta) = \frac{|\{w^s\in s\mid \exists w^c \in c\bullet\lVert w^s-w^c\rVert\leq\theta\}|}{|s|}
\end{equation}

\end{document}